\documentclass[lettersize,journal]{IEEEtran}
\usepackage{amsmath,amsfonts}
\usepackage{algorithmic}
\usepackage{algorithm}
\usepackage{array}
\usepackage[caption=false,font=normalsize,labelfont=sf,textfont=sf]{subfig}
\usepackage{textcomp}
\usepackage{stfloats}
\usepackage{url}
\usepackage{verbatim}
\usepackage{graphicx}
\usepackage{cite}
\usepackage{booktabs}
\usepackage{soul, color}
\begin{document}

\title{Disentangling segmental and prosodic factors \\to non-native speech comprehensibility}

\author{Waris Quamer, and Ricardo Gutierrez-Osuna,~\IEEEmembership{Senior Member,~IEEE}
\thanks{Waris Quamer and Ricardo Gutierrez-Osuna are with the Department of
Computer Science and Engineering, Texas A\&M University, College
Station, TX 77843 USA (e-mail: quamer.waris@tamu.edu; rgutier@tamu.edu).}
\thanks{This work was funded by NSF award 1619212 and 1623750.}}

\markboth{Submitted to IEEE/ACM TRANSACTIONS ON AUDIO, SPEECH, AND LANGUAGE PROCESSING}%
 {Shell \MakeLowercase{\textit{et al.}}: A Sample Article Using IEEEtran.cls for IEEE Journals}


\maketitle

\begin{abstract}
Current accent conversion (AC) systems do not disentangle the two main sources of non-native accent: segmental and prosodic characteristics. Being able to manipulate a non-native speaker’s segmental and/or prosodic channels independently is critical to quantify how these two channels contribute to speech comprehensibility and social attitudes. We present an AC system that not only decouples voice quality from accent, but also disentangles the latter into its segmental and prosodic characteristics. The system is able to generate accent conversions that combine (1) the segmental characteristics from a source utterance, (2) the voice characteristics from a target utterance, and (3) the prosody of a reference utterance. We show that vector quantization of acoustic embeddings and removal of consecutive duplicated codewords allows the system to transfer prosody and improve voice similarity. We conduct perceptual listening tests to quantify the individual contributions of segmental features and prosody on the perceived comprehensibility of non-native speech. Our results indicate that, contrary to prior research in non-native speech, segmental features have a larger impact on comprehensibility than prosody. The proposed AC system may also be used to study how segmental and prosody cues affect social attitudes towards non-native speech. 
\end{abstract}


\section{Introduction}
\label{sec:intro}
\IEEEPARstart{O}{lder} learners of a second language (L2) often speak with a so-called foreign accent. Unlike other aspects to L2 learning (e.g., vocabulary, grammar, writing), which can be acquired well into adulthood, achieving native-like pronunciation is difficult past a critical period because of the neuro-musculatory basis of speech production \cite{lantolf1990time}. Further, while a native accent is not required to be intelligible, improving pronunciation can reduce comprehensibility (i.e., listening effort) \cite{van2014listening} as well as social evaluations \cite{munro2003primer}. Thus, improving one’s pronunciation in an L2 offers benefits beyond intelligibility.  Several studies have suggested that practicing pronunciation with a ''golden speaker'' whose voice is similar to the L2 learner's voice  \cite{probst2002enhancing}, if not their own voice transformed to sound native-like (i.e., self-imitation training) \cite{hirose2003pronunciation}. In fact, several accent conversion (AC) techniques have been proposed for this purpose, borrowing models from the voice conversion (VC) and text-to-speech (TTS) synthesis literature \cite{wang2021accent}. AC provides a finer-grained separation of speaker characteristics than VC \cite{sisman2020overview}, since AC views accent and voice quality as independent factors to be disentangled. At present, however, AC techniques do not attempt to disentangle the two main sources of non-native accentedness: segmental and prosodic characteristics. Being able to manipulate them independently is critical to quantify how these two channels contribute to speech intelligibility, comprehensibility and social attitudes.  
This knowledge would inform the development of computer assisted pronunciation training applications based on self-imitation training \cite{ding2019golden}.

As a first step towards addressing this issue, we present a VC/AC model that provides independent control of voice quality/timbre, segmental cues and prosody\footnote{An initial version of the VC/AC model was accepted for publication at Interspeech 23. This manuscript builds on that prior work to  quantify the effect of segmental and prosodic factors on non-native speech comprehensibility (full description of the model in section III; sections IV.D, V.B, and VI are new)}. 
The model is able to generate a new utterance by combining (1) the segmental properties of an U1 utterance from any source speaker with (2) the voice quality from any target utterance U2 --as in voice cloning \cite{arik2018neural}, and (3) the prosody from a reference utterance U3 --as in expressive TTS synthesis \cite{skerry2018towards}. To achieve this goal, our model passes U1 through an acoustic model to generate a speaker-independent phonetic posteriorgram (PPG), and then to a sequence-to-sequence (seq2seq) model that combines the PPG with a speaker embedding from U2 and a prosody embedding from U3. However, naïve application of this strategy leads the seq2seq model to preserve the prosodic content in U1, which is readily available in the PPG (\textit{e.g.,} duration), instead of that in U2, which is heavily encoded. To solve this problem, we propose a technique that reduces prosodic information in the PPG, bringing it close to the information available in a phonetic transcription. Namely, we apply vector quantization (VQ) to the PPGs, and then remove consecutive duplicates. This simple trick forces the seq2seq model to use the prosodic embedding in U3 to reconstruct the speech signal. We evaluate the approach using objective and subjective measures of acoustic quality, speaker transfer and prosody transfer, and compare it against a baseline system that does not use VQ. Our results show that the proposed system achieves significantly better transfer of prosody characteristics and, as a side benefit, improved transfer of voice characteristics.

We use our proposed system to quantify the effect of segmental features and prosody on the perceived comprehensibility of accented speech through perceptual listening tests. First, we show that our system generates speech that retains the relative comprehensibility of the original utterances. Then, we assess the impact of non-native segmental and prosodic characteristics on comprehensibility by synthesizing speech with varying combinations of voice quality, segmental features, and prosody from non-native and native speaker utterances.

\section{Background and Related work}
Recent research on prosody modeling can be broadly divided into two categories, depending on whether it is being used for TTS or voice conversion (VC). In TTS, the goal of prosody modeling is to convey the desired emotions and nuances from a speaker, rather than merely convert words into sounds with the speaker’s voice quality/timbre. Most TTS systems encode prosodic information into a fixed-length embedding vector and then use it to condition the input text representation to synthesize speech. The prosody embedding is jointly learnt through an additional encoder to a seq2seq model that transforms text sequences into speech signals. For example, Skerry-Ryan et al. \cite{skerry2018towards} introduced an encoder module in a Tacotron-based \cite{wang2017tacotron} speech synthesizer to learn a prosody embedding and conditioned the synthesizer on this learnt embedding. By doing so, their system was able to synthesize audio that matched the prosody of the reference signal. Wang et al. \cite{wang2018style} introduced “global style tokens”, a bank of embeddings jointly trained with the speech synthesizer. These embeddings were trained in a self-supervised manner without any explicit labels and can be used to alter the speed of the speech signal, control, and transfer the speaking style, independently from the text content. Other methods for incorporating styles include variation inference \cite{kim2021conditional, lee2020bidirectional, zhang2019learning, hsu2018hierarchical}, flow-based modeling \cite{kim2020glow, valle2020flowtron}, and controlling pitch, duration, and energy \cite{ren2020fastspeech,valle2020mellotron}. An alternative approach to concatenating style vectors (prosody embeddings) and phoneme (or text/speech) embeddings as input to the decoder, is to introduce style through conditional normalization, such as adaptive instance normalization (AdaIN) \cite{huang2017arbitrary}. AdaIN has been applied in various speech synthesis applications, including voice conversion \cite{li2021starganv2,chen2021again} and speaker adaptation \cite{min2021meta,chen2021adaspeech}.

In VC, prosody modeling aims to alter the rhythm, melody, and intonation patterns of a person's speech while retaining their linguistic content and voice characteristics. In contrast to TTS systems --where text input is void of any prosody information, VC systems work on speech inputs where the prosody information is entangled with the segmentals and voice characteristics. To be able to alter the prosody information, VC systems need to first decouple the prosodic channel from other speech attributes. To disentangle prosody, recent approaches rely on information bottlenecks, a technique that encourages the model to learn a compressed representation of the input data to control the flow of information \cite{qian2019autovc,qian2020unsupervised}. The principle behind the approach is to force a neural network to prioritize passing information through the bottleneck that is not available elsewhere. One such system, SpeechFlow \cite{qian2020unsupervised}, used three encoder channels, each with a different information bottleneck design, and added randomly sampled noise to disentangle content, pitch, rhythm, and speaker identity. One of the major drawbacks of this approach is that bottlenecks need to be carefully designed, as they are sensitive to the dimensionality of latent space. Studies have employed vector quantization (VQ) to force bottlenecks, by compressing the continuous speech representation into a small number of discrete clusters.  VQVC \cite{wu2020one} used VQ to disentangle speaker and content representations and trained the model to reconstruct the speech signal. VQVC+ \cite{wu2020vqvc+} improved VQVC performance using a U-net architecture and an auto-encoder based system to generate audio of high quality. However, the content and speaker representations in these systems may still be entangled. An alternative approach to force disentanglement between various speech attributes is to use the mutual information (MI) loss \cite{wang21n_interspeech}. The MI loss penalizes the similarity between different latent representations and allows more precise control over them. An example of this approach is VQMIVC \cite{wang21n_interspeech}, a model that additionally used MI loss with VQ to decorrelate content and speaker representation. Other approaches include a modified Variational Auto-Encoder (VAE) loss function that penalizes the similarity between different attribute representations \cite{kumar2021learning}, and a self-supervised contrastive model that uses product quantization to disentangle non-timbral prosodic information from raw audio \cite{weston2021learning}.

\section{Methods}
\label{sec:method}
\begin{figure}
    \centering
    \includegraphics[scale=0.7]{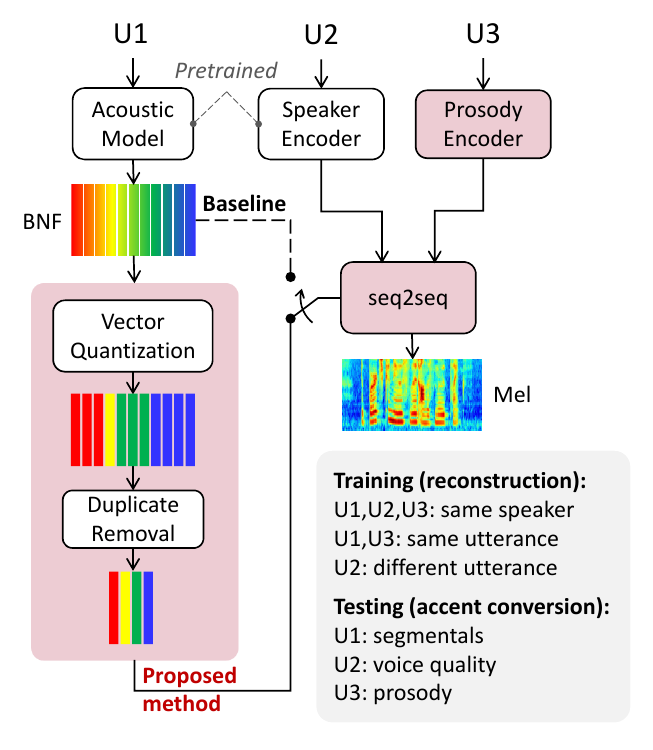}
    \vspace{-5mm}
    \caption{Block diagram of the proposed system. The prosody encoder and seq2seq model are trained jointly as an auto-encoder. For \textit{accent conversion}, segmentals come from U1 and  prosody from U3, thus providing independent control of both channels}.
    \label{fig:block_diagram}
\vspace{-10pt}
\end{figure}

We propose a voice/accent conversion system that can decouple segmentals from prosodic characteristics, thus providing independent control of these channels when synthesizing new speech signals. The proposed model is illustrated in Figure \ref{fig:block_diagram}. An input utterance (U1) is passed through an acoustic model (AM) to generate a bottleneck feature (BNF) matrix that captures the phonetic content of the utterance –see section \ref{sec:acoustic_model}. The BNF matrix is then passed to (1) a vector quantization (VQ) module that discretizes each column (i.e., frame) into one of N codewords (i.e., cluster centers), and then (2) a duplicate removal (DR) stage that eliminates consecutive duplicates of each codeword. The resulting short sequence of codewords can be viewed as a sequence of phonemic codes (for N=39) or sub-phonemic codes (for larger N). Thus, it is akin to a phonemic transcription, except phonemes are not represented by symbols but by their corresponding BNF codewords.

A seq2seq model consumes (1) the short sequence of BNF codewords from utterance U1, (2) a speaker embedding representing the voice quality in utterance U2 from a target speaker, and (3) a prosody embedding from a reference utterance U3. From these three information bottlenecks, the seq2seq attempts to reconstruct the original Mel spectrogram. The prosody encoder and seq2seq model are trained simultaneously in an unsupervised fashion (i.e., as an auto-encoder) while the speaker encoder and acoustic model are pre-trained in advance. During training, the AM and prosody encoder are fed the same utterance from the same speaker, whereas the speaker embedding is fed a different utterance U2 from the same speaker. This trick ensures that (1) the prosody encoder learns a different mapping than the speaker encoder, and that (2) the seq2seq model does not attempt to infer prosody from the speaker embedding.

\subsection{Acoustic Model (AM)}
\label{sec:acoustic_model}
The AM serves the function of generating a speaker-independent speech representation. The AM uses a Factorized Time Delayed Neural Network (TDNN-F) \cite{povey2018semi} to generate a phonetic posteriorgram (PPG) that represents the posterior probability of each speech frame belonging to predefined phonetic units (\textit{e.g.,} phonemes or triphones/senones). The TDNN-F architecture consists of five hidden layers with Rectified Linear Unit (ReLU) activation, culminating in a layer with 256 neurons \cite{peddinti2015time}. The AM is trained on the Librispeech corpus \cite{panayotov2015librispeech}, which contains recordings from thousands of native speakers. Following Dehak \textit{et al.} \cite{dehak2010front}, the input to the model consists of an acoustic feature vector (40-dim MFCC) combined with an i-vector (100-dim) representing the speaker. In line with our previous work \cite{quamer2022zero}, we extract BNFs from the AM model. These BNFs are derived from the last hidden layer, prior to the final softmax layer. Compared to the higher-dimensional PPGs generated by the final softmax layer (6,024 for senone-PPGs), BNFs offer a significantly reduced dimensionality of 256 \cite{zhao2021converting}. This dimensionality reduction greatly simplifies the subsequent training process of the seq2seq model, while still capturing essential linguistic information. By leveraging the advantages of BNFs, we benefit from reduced computational complexity and memory requirements.

\subsection{Speaker Encoder}
To capture the voice quality of each speaker, we use a speaker encoder trained as a speaker-verification model, following the framework outlined in \cite{wan2018generalized}.  The speaker encoder generates a fixed-dimension embedding vector that captures the unique acoustic characteristics of each speaker. The architecture of our speaker encoder consists of a 3-layer LSTM with 256 hidden nodes per layer. The final LSTM layer's hidden state is then passed through a projection layer with 256 units. During training, we employ the generalized end-to-end (GE2E) loss function \cite{wan2018generalized}, which maximizes the cosine similarity between utterances from the same speaker. This loss function trains the speaker encoder to effectively discriminate and differentiate between different speakers. 

\subsection{Sequence-to-Sequence Model}
Our seq2seq model is derived from the Voice Transformer Network \cite{li2019neural}, which is a combination of Transformer \cite{vaswani2017attention} and Tacotron2 \cite{shen2018natural}. Following Li \textit{et al.} \cite{li2019neural}, we adapt the Transformer architecture to match the VC task by adding pre-nets to the decoder. We add an extra linear layer with a weighted binary cross-entropy loss to predict the stop token. Similar to Tacotron models \cite{shen2018natural}, we used a five-layer CNN postnet to predict a residual that refines the final prediction. The detailed architecture and model parameters are presented in Table \ref{tab:params_seq2seq}.

The seq2seq model consumes a BNF matrix and outputs a converted log-Mel spectrogram. The high time resolution of both input and output acoustic features in VC makes attention learning difficult, and increases the training memory footprint. While training our baseline model, we use a reduction factor $r_e$ and $r_d$ on both encoder and decoder side, respectively, so that it can stack multiple frames to reduce the time axis. This not only improves attention alignment but also reduces the training memory footprint by half, as well as the number of required gradient accumulation steps \cite{Huang2020}. When using duplicate removal, the time resolution of the input vector quantized BNFs is comparable to text inputs in TTS systems, and much lower than those of original acoustic features. So, in the latter case, we employ the reduction factor $r_d$ only on the decoder side.

\begin{table}[]
  \caption{Parameters of the seq2seq architecture}
  \label{tab:params_seq2seq}
  \centering
  \resizebox{\linewidth}{!}{%
  \begin{tabular}{ll}
    \toprule
    \textbf{Components}      & \textbf{Parameters}        \\
    \midrule
    Input Dimension & 256 (BNFs) \\
    Speaker Embedding & 256D \\
    Prosody Embedding & 256D \\
    \midrule
    Encoder	        & 4-layer Transformer, 512D attention layer, \\
                    & 1024 dense units \\
                    & reduction factor ($r_e$) is 1 for proposed,\\ &2 for baseline  \\
    \midrule
    Decoder PreNet	& Two FC layers, each has 256 ReLU units, \\
                    & 0.5 dropout rate \\
    Decoder	        & 4-layer Transformer, 512D attention layer, \\
                    & 1024 dense units \\
                    & reduction factor ($r_d$) = 2 \\
    Decoder PostNet	& Five 1D convolution layers (kernel size 5), \\
                    & 0.5 dropout rate, 512 channels in the first \\
                    & four layers and 80 channels in the last layer\\
    \midrule
    Output dimension&80D Mel-spectrogram\\

    \bottomrule
  \end{tabular}}
\vspace{-10pt}
\end{table}

\subsection{Prosody Encoder}
Our prosody encoder is based on the ECAPA-TDNN model \cite{desplanques2020ecapa}, and is designed to extract a vector embedding that compresses the prosodic information in utterance U3. The model architecture begins with a TDNN layer, which captures temporal dependencies in the input features. This is followed by three SE-Res2Blocks, each comprising two 1D-CNN layers, a dilated Res2Net, and a Squeeze-Excitation (SE) block. The outputs from the SE-Res2Blocks are then combined using a 1D-CNN layer. Subsequently, we apply attentive statistics pooling to the extracted features. Finally, a Fully Connected (FC) layer produces a prosody embedding, which compresses the prosodic characteristics into a 256-dimensional vector.

In the first three SE-Res2Blocks, we use dilation factors of 2, 3, and 4, respectively. The channel size for these blocks is set to 1024, with a kernel size of 3. It is important to note that during training the prosody encoder receives the Mel-spectrogram from the same utterance as the one fed to the AM. This ensures consistent information processing and alignment between the acoustic and prosodic features. We train the prosody encoder jointly with the seq2seq model. During training, the prosody encoder attempts to minimize the mean square error between the generated and original Mel-spectrograms. This joint training approach allows for the seamless re-integration of prosodic cues with the acoustic features in the synthesis process.


\subsection{Vector Quantization and Duplicate removal}
Though the BNF matrix primarily captures segmental information in U1, it also preserves significant prosodic characteristics (\textit{e.g.,} phone duration, speaking rate). As such, were the BNF matrix to be used as an input, the seq2seq model would have to learn to ignore its prosody content (which is that of U1) and instead focus on the prosody embedding from utterance U3. However, prosody in U1 is trivially available (i.e., the number of columns in the BNF matrix equals the duration), whereas prosody in U3 is encoded into a compact vector. As such, the seq2seq will generally converge to a local minimum that ignores the prosody encoding and instead preserves the prosody in the BNF matrix.

To avoid this local minimum, we propose to remove prosodic content in the BNF matrix using vector quantization (VQ). Namely, we pre-train a k-means model to learn a set of codewords (\textit{i.e.,} cluster centers) from the L2-ARCTIC corpus \cite{zhao2018l2} (20 speakers, 1,000 utterances each). Once the codebook has been learned, we replace each column in the BNF matrix with its corresponding codeword, and finally eliminate any duplicate codewords that are adjacent in the sequence, as depicted in Figure \ref{fig:block_diagram}. In this fashion, timing information is removed from the BNF matrix, which is reduced to a short sequence of codewords that only preserves key segmental information in U1.

When we use this short codebook sequence to jointly train the prosody encoder and the seq2seq model, the two modules are forced to learn complementary tasks. The prosody encoder is forced to learn to generate an embedding that summarizes the prosody in U3. And, in turn, the seq2seq model is forced to learn to combine the prosody embedding with the short codebook sequence to reconstruct the original Mel-spectrogram. Though not our main focus, a second major advantage of vector quantizing BNFs is that it can lead to significant improvements in voice conversion performance, as shown in prior studies \cite{wu2020vqvc+}. It is important to note that this secondary benefit is due to the VQ step alone, not the subsequent DR step.

\subsection{Experimental Setup}
We developed the AM using the Kaldi framework \cite{povey2011kaldi} and trained it on the Librispeech corpus \cite{panayotov2015librispeech}, which includes recordings from 2,484 native speakers of American English. The trained AM achieves a word error rate (WER) of 3.76\% on the test-clean subset of Librispeech. To train the speaker encoder, we combined data from VoxCeleb1 \cite{nagrani2020voxceleb}, VoxCeleb2 \cite{chung2018voxceleb2}, and Librispeech, resulting in approximately 3,000 hours of speech data from 9,847 speakers. The speaker encoder is implemented using PyTorch and trained using the Adam optimizer with a learning rate of 10-3 and a batch size of 128. We trained our seq2seq model and the prosody encoder using the ARCTIC \cite{kominek2004cmu} and L2-ARCTIC \cite{zhao2018l2} corpora. The combined dataset consists of 28 speakers (1,132 utterance each). From these, we excluded four speakers (NJS, YKWK, TXHC and ZHAA) from L2-ARCTIC and use them as unseen speakers during testing. For all speakers, we further divided their utterances into three non-overlapping subsets: a training set of 1,032 utterances, a validation set of 50 utterances and a test set of 50 utterances. The seq2seq model consisted of 4 encoding layers and 4 decoding layers, both with reduction factors $r_e$ and $r_d$ of 2. We set the batch size to 16 and used the AdamW optimizer with a learning rate of $10^{-3}$ annealed down to $10^{-5}$ by exponential scheduling. We used the k-means algorithm to generate the VQ model and trained it on the same training split as the seq2seq. To convert Mel-spectrograms to waveforms, we used a pre-trained HiFiGAN vocoder \cite{kong2020hifi}. For all the models, we extracted 80-dim Mel-spectrograms with 25ms window and 10ms shift. All our models were trained using two NVIDIA Tesla V100 GPUs. We use speaker BDL from ARCTIC \cite{kominek2004cmu} as the reference L1 speaker for system evaluation experiments. We conduct our comprehensibility experiments using two native speakers BDL (male, American) and RMS (male, American) and two non-native speakers TLV (male, Vietnamese) and EBVS (male, Spanish).

\section{Results}
In this section, we present results from two sets of experiments. In the first set of experiments, we evaluated the synthesis quality of our proposed system against a baseline system that did not use vector quantization and duplicate removal --see Figure \ref{fig:block_diagram}. In the second set of experiments, we used our proposed system to quantify the role of segmentals and prosody on comprehensibility.


\subsection{Synthesis quality}
\begin{figure}[]
    \centering
    \includegraphics[scale=0.65]{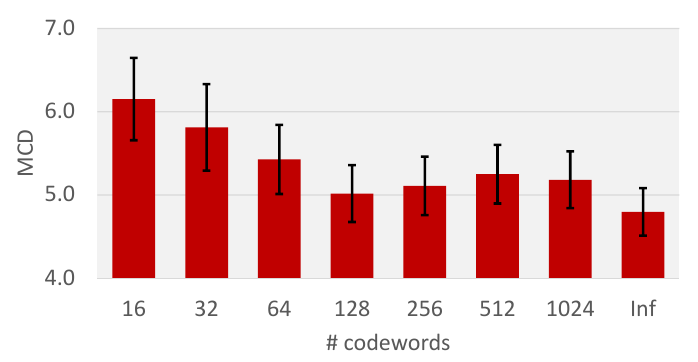}
    \vspace{-8pt}
    \caption{Mel Cepstral Distortion (MCD) vs. codebook size.  The lowest MCD is reached when the number of codewords is infinite (baseline system). MCD decreases as the number of codewords increases, and stabilizes after 128 codewords}    
    \label{fig:MCD}
    \vspace{-8pt}
\end{figure}

In a first step, we evaluated synthesis quality using objective and subjective measures. As an objective measure, we examined how the size of the codebook impacted Mel Cepstral Distortion (MCD). For this purpose, we used the proposed system as an auto-encoder: to reconstruct at the output the same utterance fed to the acoustic model, prosody encoder and speaker encoding (\textit{i.e.,} $U1=U2=U3$). We performed this experiment on four L2-ARCTIC speakers (NJS, YKWK, TXHC and ZHAA) that had been held out when training the prosody embedding and seq2seq model. Results are shown in Figure \ref{fig:MCD} for different codebook sizes; the baseline system (i.e., without VQ) is equivalent to having an infinite number of codewords ($vq\infty$).

As shown, MCD decreases significantly as the codebook size increases up to 128 codewords, and then stabilizes. One-way ANOVA shows that the effect of codebook size is statistically significant $F(7, 16) = 14.23, p\ll0.001$. Further, paired t-test shows a significant difference between $vq128$ and $vq64$ $(p = 0.007, one-tailed)$, and between $vq128$ and $vq\infty$ $(p = 0.005, one-tailed)$. Thus, while the lowest MCD is achieved when the VQ-DR step is not used (the baseline system, $vq\infty$), the lowest MCD among all the VQ models is for $128$ codewords. As such, all subsequent models in this study are based on $vq128$.

To verify these results perceptually, we conducted listening tests on Amazon Mechanical Turk (AMT), where listeners (N=20) were asked to rate the acoustic quality of utterances. Following \cite{quamer2022zero}, we used a standard 5-point scale mean opinion score (MOS) as follows [rating, speech quality, level of distortion]: [5, excellent, imperceptible] — [4, good, just perceptible but not annoying] — [3, fair, perceptible but slightly annoying] — [2, poor, annoying but not objectionable ] — [1, bad, very annoying and objectionable]. Each listener rated 20 utterances from the $vq128$ model (proposed) and the $vq\infty$ model (which served as a baseline), as well as original L2 utterances. Results are summarized in Table \ref{tab:MOS}. As expected, the original L2 utterances received the highest MOS ratings $(4.32)$. Speech quality dropped by $0.43$ MOS points ($p\ll0.001$) for the baseline system, and an additional 0.20 points ($p\ll0.001$) for the proposed system. While this result was also expected (and consistent with the objective results in Figure \ref{fig:MCD}), it is noteworthy that discretizing the speech spectrum down to 128 codewords achieves nearly the same synthesis quality as using the full range of spectral variability in the speech corpus.

\begin{table}[]
\footnotesize
  \caption{MOS for baseline ($vq\infty$) and proposed ($vq128$) systems}
  \label{tab:MOS}
  \centering
  \begin{tabular}{lllll}
\toprule
    & \textbf{Target} & \textbf{Baseline} & \textbf{Proposed} & \textbf{p value}                    \\ \midrule
MOS & 4.32 $\pm$ 0.82  & 3.89  $\pm$ 0.83 & 3.69 $\pm$ 0.79   & $\ll$ 0.001 \\ \bottomrule
\end{tabular}
\vspace{-10pt}
\end{table}

\subsection{Transfer of speaker identity}
\begin{figure}[]
    \centering
    \includegraphics[scale=0.65]{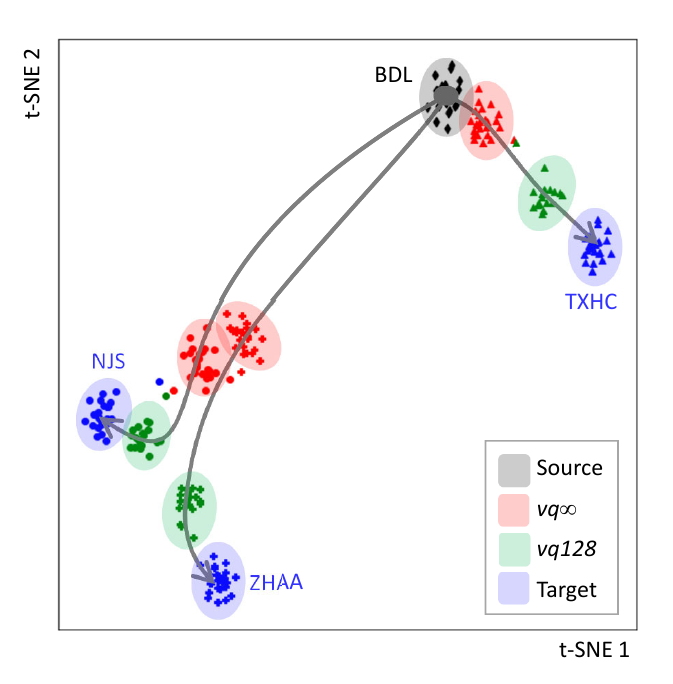}
    \vspace{-5mm}
    \caption{t-SNE of speaker embeddings 
for source (black), target (blue), baseline $vq\infty$ (red) and proposed $vq128$ (green).  The arrows represent a path connecting source and target utterances, passing through conversions from the two systems. Conversions from $vq128$ are much closer to the target than those from $vq\infty$, indicating that the $vq128$ system provides better transfer of speaker identity. }    
    \label{fig:embedding}
\vspace{-10pt}
\end{figure}

As we had done to evaluate synthesis quality, we used objective and subjective measures to evaluate speaker transfer in models $vq\infty$ (baseline) and $vq128$ (proposed). As an objective measure, we visualized the embeddings produced by the speaker encoder for the source speaker (BDL), three target speakers (NJS, TXHC, ZHAA), and voice conversions from both systems, 10 utterances per voice. Results are shown as a tSNE plot in Figure \ref{fig:embedding}. We find that speaker transfer is inversely related to the distance between each voice conversion ($vq\infty$, $vq128$) and the corresponding target speaker. As shown, voice conversions from $vq128$ are significantly closer to their target than those from $vq\infty$, indicating that the VQ-DR step improves transfer from source to target speaker.

To corroborate these results, we conducted ABX listening tests on AMT, where participants (N=20) were presented with two audio samples, one from $vq\infty$ and one from $vq128$ (in a counterbalanced fashion), followed by the original L2 utterance. Then, participants had to decide which audio sample ($vq\infty$ or vq128) was closest the L2 utterance in terms of voice/timbre, and then rate the confidence in their decision using a 7-point scale (7: extremely confident; 5: quite a bit confident; 3: somewhat confident; 1: not confident at all). Participants were instructed to focus only on the voice and ignore any noises or distortions. Following \cite{felps2010developing}, the decision and confidence levels were then collapsed to form a 14-point VSS (Voice Similarity Score) scale: -7 (definitely $vq\infty$) to +7 (definitely $vq128$). Each listener rated 10 ABX triplets per L2 speaker and system. As shown in Table \ref{tab:speaker similarity}, listeners chose $vq128$ outputs as the closest to the L2 speaker 70.25\% of the times, and with a high confidence level (5.12: quite a bit confident it is $vq128$ ), whereas the baseline ($vq\infty$) was selected only 29.75\% of the times, and with a low confidence level (1:22: not confident at all it is $vq\infty$.) This result further corroborates the qualitative results in the t-SNE plot in Figure \ref{fig:embedding}.

\begin{table}[]
  \caption{Perceptual ratings of \textbf{speaker transfer} in an ABX test}
  \label{tab:speaker similarity}
  \centering
  \footnotesize
  \resizebox{\linewidth}{!}{%
  \begin{tabular}{lll}
    \toprule
    \textbf{Rating}      & \textbf{Baseline ($vq\infty$)}    &\textbf{Proposed ($vq128$)}        \\
    \midrule
    Closest to the L2 speaker & 29.75\%  & 70.25\%  \\
    Average rater confidence & 1.22  & 5.12  \\
    \bottomrule
  \end{tabular}}
\end{table}

\subsection{Prosody transfer}
We also examined how well the $vq\infty$ (baseline) and $vq128$ (proposed) models were able to transfer the prosodic characteristics of utterance U3. For this purpose, utterance U1 was from an L1 speaker, whereas utterances $U2=U3$ were from an L2 speaker. As such, the system was expected to generate an utterance with L1 segmentals and L2 prosody. For objective evaluation, we measured differences in duration, average F0 and F0 range between conversions from both systems and utterances U1/L1 (i.e., whose prosody should be ignored) and U2/L2 (i.e., whose prosody should be transferred). If prosody transfer was successful, we hypothesized that the duration, F0 average and F0 range for the voice conversions would be closer to those of the L2 utterance than to those in the L1 utterance. Results in Table \ref{tab:prosodic differences} confirm this hypothesis for the $vq128$ system, but the reverse hypothesis for the $vq\infty$ (baseline) system. Namely, the three measures of prosody for $vq\infty$ syntheses are closer to the L1 utterance, whereas for the $vq128$ system the three measures are closer to the L2 utterance. Thus, these results indicate that only the $vq128$ system is able to transfer the prosody characteristics present in the reference utterance U3.

\begin{table}[]
\caption{Differences in prosodic characteristics between original utterances (L1, L2) and accent conversions ($vq\infty$, $vq128$)}
\label{tab:prosodic differences}
\footnotesize
\resizebox{\linewidth}{!}{%
\begin{tabular}{c cc cc cc}
\hline
       & \multicolumn{2}{c}{\begin{tabular}[c]{@{}c@{}}\textbf{$\Delta$ duration}\\ (ms)\end{tabular}} & \multicolumn{2}{c}{\begin{tabular}[c]{@{}c@{}}\textbf{$\Delta$ F0 avg}\\ (Hz)\end{tabular}} & \multicolumn{2}{c}{\begin{tabular}[c]{@{}c@{}}\textbf{$\Delta$ F0 range}\\ (Hz)\end{tabular}} \\ \hline
       & \multicolumn{1}{c}{L1}                         & L2                         & \multicolumn{1}{c}{L1}                          & L2                          & \multicolumn{1}{c}{L1}                         & L2                         \\ \hline
$vq\infty$     & \multicolumn{1}{c}{\textbf{16.89} }                    &         413.37                   & \multicolumn{1}{c}{\textbf{36.83}}                            &         40.43                    & \multicolumn{1}{c}{\textbf{48.53}}                           &           35.93                 \\ \hline
$vq128$ & \multicolumn{1}{c}{395.42}                           &        \textbf{ 5.89}                   & \multicolumn{1}{c}{82.36}                            &          \textbf{ 7.96}                  & \multicolumn{1}{c}{20.07}                           &              \textbf{11.42}              \\ \hline
\end{tabular}}
\end{table}

To corroborate these findings perceptually, we conducted a second ABX test on AMT, where participants (N=20 listened to audio samples from both systems ($vq\infty$ or $vq128$) in a counterbalanced fashion, followed by the original L2 utterance. As before, participants had to decide which audio sample ($vq\infty$ or $vq128$) was closest to the L2 utterance in terms of the speaking style, and then rate their confidence. Participants were instructed to focus on prosody specific attributes of the recordings such as speaking rate, pauses and intonations and ignore the audio quality (e.g., noises, distortions). Results are shown in Table \ref{tab:speaking style similarity}. Listeners rated utterances from the proposed system ($vq128$) as the closest to the original L2 utterance 69\% of the times with somewhat confidence (3.2), whereas the $vq\infty$ was selected the remaining 31.12\% of the times with very low confidence (1.2). This result is remarkable considering that listeners were instructed to focus on speaking style rather than differences in segmental content between the two accent conversions (L1 segmentals) and the L2 utterances (L2 segmentals).

\begin{table}[]
  \caption{Perceptual ratings of \textbf{prosody transfer} in an ABX test}
  \label{tab:speaking style similarity}
  \centering
  \footnotesize
  \resizebox{\linewidth}{!}{%
  \begin{tabular}{lll}
    \toprule
    \textbf{Rating}      & \textbf{Baseline ($vq\infty$)}    &\textbf{Proposed ($vq128$)}        \\
    \midrule
    Closest to the L2 speaker & 31.12\% & 68.88\% \\
    Average rater confidence & 1.31 & 3.2   \\
    \bottomrule
  \end{tabular}}
\end{table}

\subsection{Quantifying the role of segmentals and prosody on comprehensibility}
\label{sec:quantify}
Having established the validity of the voice/accent conversion system, we can now focus on the overarching question that motivated this study: understanding the impact of segmental features and prosody on the perceived comprehensibility of accented speech through perceptual listening tests. In a first experiment, we establish that the L2 utterances are less comprehensible than the L1 utterances --otherwise, any subsequent experiment would be inconclusive. In a second experiment, and for the same reason, we establish that our proposed accent conversion method does preserve the relative comprehensibility of the original L1 and L2 utterances. On the remaining experiments, we use our system to perform multi-accent voice conversions, \textit{i.e.,} converting a non-native voice to a native voice while retaining non-native segmental and prosody, and vice-versa), and used them to evaluate the effect of voice quality on comprehensibility. Finally, we evaluate the independent effects of non-native segmental and prosodic cues on comprehensibility. The results and experiment design are described in detail below.

\textbf{Listening tests.} We conducted listening tests on AMT with participants who self-reported English was their native language. In addition, participants were required to reside in the United States and had to pass a screening test that asked them to identify regional dialects of American English \cite{aryal2013foreign}: Northeast (\textit{i.e.,} Boston, New York), Southern (\textit{i.e.,} Georgia, Texas, Louisiana), and General American (\textit{i.e.,} Indiana, Iowa). Each experiment presented participants with pairs of audio recordings, each featuring a unique combination of voice identity, segmental features, and prosodic properties. Participants were asked to listen to 16 pairs of recordings and then select the recording that required the least effort to understand (i.e., comprehensibility). Participants were instructed to focus on the words being uttered by the speaker, and ignore noise or distortions in the audio. They also rated their confidence in their decision (CR) using a 7-point scale (7: extremely confident, 1: not confident at all). The pairs were randomized to avoid order effects.

\textbf{Notation.} For clarity, we denote a speech sample as a triplet $\{Q_i,S_j,P_k\}$, where Q represents the voice quality of the speaker, S represents the segmental properties of the utterance, P represents its prosodic properties, and the subindices $\{i,j,k\}\in(1,2)$ indicate that the corresponding attributes is modeled from a native speaker of English (L1) or a non-native speaker (L2). For instance, $\{Q_2,S_1,P_2\}$ signifies synthetic speech with the voice quality and prosody of an L2 speaker, and the segmental attributes of an L1 speaker.
\vspace{2mm}
\subsubsection{\ul{Comprehensibility of \textbf{original} L1 and L2 utterances}}
In the first experiment, we asked listeners (N=20)  to evaluate the comprehensibility of original (i.e., not synthesized) utterances from the L1 and L2 speakers.  This allowed us to establish that the L2 utterances were perceived as less comprehensible than the L1 utterances, before we manipulated them through synthesis.  For this purpose, listeners were presented with pairs of audio recordings, each consisting of an original recording from a non-native speaker (L2) paired with a parallel utterance from a native speaker (L1), and were asked to select the utterance which took them less effort to understand.

Results are summarized in Table \ref{tab:compre_original}, and indicate L1 utterances were perceived as more comprehensible than L2 utterances 89.1\% of the instances and with a higher confidence rating (6.4). These results align with current literature \cite{munro1995foreign} and confirm that the L2 speakers in our corpus did have lower comprehensibility than the L1 speakers. 

\begin{table}[]
  \caption{Perceptual ratings of \textbf{original} L1 vs L2 in an AB test}
  \footnotesize
  \label{tab:compre_original}
  \centering
  \begin{tabular}{lll}
    \toprule
          & \textbf{Original L1 speech}    &\textbf{Original L2 speech}        \\
    \midrule
    Choice (\%) & \textbf{89.1\%} & 10.9\% \\
    Average CR & 6.4 & 5.5   \\
    \bottomrule
  \end{tabular}
\end{table}

\vspace{2mm}
\subsubsection{\ul{Comprehensibility of \textbf{reconstructed} L1 and L2 speech}}
In the second experiment, we sought to validate that our proposed accent conversion system could preserve the relative comprehensibility of L1 and L2 speech. For this purpose, we use the system as an auto-encoder to reconstruct at the output the same utterance that was fed the acoustic model, the speaker encoder and the prosody encoder (i.e., $U1=U2=U3$).  The experiment involved reconstructing the same pairs of utterances from the L1 speaker $\{Q_1,S_1,P_1\}$ and L2 speakers $\{Q_2,S_2,P_2\}$ of the previous experiment, but in this case after they were passed through our system. Participants (N=20) were then presented with pairs of synthesized utterances and were asked to choose the utterance which took them less effort to understand and rate their confidence.

Results are summarized in Table \ref{tab:compre_recons} and are virtually identical to those on original L1 and L2 utterances (Table \ref{tab:compre_original}). Namely, resynthesized L1 utterances $\{Q_1,S_1,P_1\}$ were rated as more comprehensible than resynthesized L2 utterances $\{Q_2,S_2,P_2\}$, indicating that our accent conversion system does preserve the segmental and prosodic cues that affect comprehensibility. 

\begin{table}[]
  \caption{Perceptual ratings of comprehensibility \\for \textbf{reconstructed} L1 and L2 speech}
  \label{tab:compre_recons}
  \centering
  \footnotesize
  \begin{tabular}{lll}
    \toprule
          & \textbf{$\{Q_1,S_1,P_1\}$}    &\textbf{$\{Q_2,S_2,P_2\}$}        \\
    \midrule
    Choice (\%) & \textbf{90.6\%} & 9.4\% \\
    Average CR & 6.3 & 5.0   \\
    \bottomrule
  \end{tabular}
\end{table}

\vspace{2mm}
\subsubsection{\ul{Effect of voice quality on comprehensibility}}
In a third experiment, we examined the effect of voice quality on speech comprehensibility to determine whether the speaker embedding accidentally captured any accent information. For this purpose, we used the proposed system to generate accent-converted speech in two directions: the voice quality of an L2 speaker with the segmental and prosodic characteristics of an L1 speaker ($\{Q_2,S_1,P_1\}$), and the reverse: the voice quality of an L1 speaker with the segmental and prosodic characteristics of an L2 speaker ($\{Q_1,S_2,P_2\}$). These accent-converted utterances were then paired with the original utterances synthesized with their original accents: $\{Q_2,S_1,P_1\}$ was paired with $\{Q_2,S_2,P_2\}$, and $\{Q_1,S_2,P_2\}$ was paired with $\{Q_1,S_1,P_1\}$. As before, participants (N=20) were presented with pairs of utterances, and were asked to choose the one which took them less effort to understand, as well as their confidence in the assessment.

The results are presented in Table \ref{tab:compreh_voice}. In the first part of the experiment (removing a non-native accent), most listeners (81.9\%) indicated that resynthesizing the voice of an L2 speaker with native segmentals and prosody $\{Q_2,S_1,P_1\}$ had higher comprehensibility than the L2 speaker’s voice with its original non-native segmentals and prosody $\{Q_2,S_2,P_2\}$. In the second part of the experiment (adding a non-native accent), most listeners (86.9\%) indicated that resynthesizing the voice of an L1 speaker with non-native segmentals and prosody $\{Q_1,S_2,P_2\}$ reduced the L1 speaker’s comprehensibility, compared to the L1 speaker’s voice with its original native segmentals and prosody  $\{Q_1,S_1,P_1\}$. Overall, irrespective of voice quality, participants chose utterances synthesized with native segmental and prosody to be more comprehensible and with a higher confidence rating. These findings indicate that listeners did not associate certain voices with their original accents. Moreover, voice quality did not affect comprehensibility, confirming that the speaker embedding of voice quality did not capture accent information.

\begin{table}[]
\caption{Perceptual ratings of comparisons \\between accent converted speech in an AB test}
\centering
\label{tab:compreh_voice}
\resizebox{\linewidth}{!}{%
\begin{tabular}{c cc cc}
\hline
       & \multicolumn{2}{c}{\begin{tabular}[c]{@{}c@{}}\textbf{Voice Quality: L2 (2)}\end{tabular}} 
       & \multicolumn{2}{c}{\begin{tabular}[c]{@{}c@{}}\textbf{Voice Quality: L1(1)}\end{tabular}} \\
\hline
       & \multicolumn{1}{c}{$\{Q_2,S_1,P_1\}$}  & $\{Q_2,S_2,P_2\}$                         
       & \multicolumn{1}{c}{$\{Q_1,S_1,P_1\}$}  & $\{Q_1,S_2,P_2\}$ \\
\hline
Choice (\%) & \multicolumn{1}{c}{\textbf{81.9\%} } & 18.1\%                  
            & \multicolumn{1}{c}{\textbf{86.9\%}} & 13.1\%  \\
Average CR & \multicolumn{1}{c}{5.8} &  4.9                   
                            & \multicolumn{1}{c}{6.1}  &   5.3  \\          
\hline
\end{tabular} }
\end{table}

\vspace{2mm}
\subsubsection{\ul{Effect of non-native segmental and prosodic cues to comprehensibility}}
In a final experiment, we sought to quantify the relative effect of segmental and prosodic cues to the comprehensibility on non-native speech. For this purpose, we used our system to generate two distinct syntheses: L2 speech modified to have native segmentals while retaining its voice quality and prosody $\{Q_2,S_1,P_2\}$, and L2 speech modified to have native prosody while retaining its voice quality and segmentals $\{Q_2,S_2,P_1\}$. As before, participants (N=20) listened to pairs of synthesized utterances, and then chose the utterance that took them less effort to understand (along with their confidence level). We repeated the same procedure with L1 utterances, generating two sets of syntheses that matched the non-native segmental $\{Q_1,S_2,P_1\}$ and prosody $\{Q_1,S_1,P_2\}$ respectively. 

The results are summarized in Table \ref{tab:compreh_nns_seg}. In the first part of the experiment (L2 speakers), most listeners (76.6\%) indicated that resynthesizing L2 speech with native segmentals improved L2 speaker’s comprehensibility $\{Q_2,S_1,P_2\}$, compared to L2 speech resynthesized with native prosody $\{Q_2,S_2,P_1\}$. In the second part of the experiment (L1 speakers), most listeners (76.6\%) found L1 speech with non-native segmental features $\{Q_1,S_2,P_1\}$ reduced L1 speaker’s comprehensibility, compared to L1 speech with non-native prosody $\{Q_1,S_1,P_2\}$. These findings suggest that segmental features have a greater impact on comprehensibility than prosody.

\begin{table}[]
\caption{Perceptual ratings of comparison between segmental and prosodic conversions in an AB test}
\centering
\label{tab:compreh_nns_seg}
\resizebox{\linewidth}{!}{%
\begin{tabular}{c cc cc}
\hline
       & \multicolumn{2}{c}{\begin{tabular}[c]{@{}c@{}}\textbf{Voice Quality: L2 (2)}\end{tabular}} 
       & \multicolumn{2}{c}{\begin{tabular}[c]{@{}c@{}}\textbf{Voice Quality: L1 (1)}\end{tabular}} \\
\hline
       & \multicolumn{1}{c}{$\{Q_2,S_1,P_2\}$}  & $\{Q_2,S_2,P_1\}$                         
       & \multicolumn{1}{c}{$\{Q_1,S_1,P_2\}$}  & $\{Q_1,S_2,P_1\}$ \\
\hline
Choice (\%) & \multicolumn{1}{c}{\textbf{71.6\%} } & 28.4\%                  
            & \multicolumn{1}{c}{\textbf{76.6\%}} & 23.4\%  \\
Average CR & \multicolumn{1}{c}{5.1} &  4.9                   
                            & \multicolumn{1}{c}{5.5}  & 4.7  \\          
\hline
\end{tabular} }
\end{table}

\section{Discussion}
We have proposed a methodology to evaluate the individual contributions of segmental and prosodic features on the perceived comprehensibility of non-native speech. The methodology combines an accent conversion system that can manipulate specific aspects of speech (i.e., voice quality, segmental, and prosody) independently to synthesize new speech signals, with an experimental protocol for perceptual listening tests. In this section, we discuss the results obtained from our experiments and point out future research directions.

\subsection{Accent conversion system}
We compared our proposed accent conversion system against a baseline that did not use vector quantization and the subsequent duplicate removal (VQ-DR) step. Both subjective and objective evaluation indicate that our proposed model significantly outperforms a baseline system without VQ-DR in speaker identity transfer while synthesizing speech with similar acoustic quality. More importantly, the proposed system is able to transfer prosodic characteristics from a reference utterance while the baseline retains the prosody of the source utterance. A possible explanation for this result is that by discretizing speech (vector quantizing phonetic content) down to a small number  of codewords, the model creates an information bottleneck that further reduces any residual speaker information in the BNF. Next, the duplicate removal step (subsampling in time) removes any duration (or rhythm) information, thus forcing the seq2seq model to reconstruct speech using the prosody of the reference speech. In the baseline implementation (without VQ-DR), duration information is trivially available in the BNFs and therefore the seq2seq models ignores information from prosody embeddings. 


The accent-conversion system synthesizes speech with “good” acoustic quality with noise “just perceptible but not annoying”. The synthesis quality (and potential speedup to the speech generation process) can be further improved by directly integrating HiFiGAN \cite{kong2020hifi} into the decoding process. This would enable the system to directly generate waveforms from the latent representation, bypassing the intermediate step of producing the Mel-spectrogram. The non-autoregressive nature of HiFiGAN as decoder would necessitate explicit prediction of codeword repetitions and pitch values. Following a similar approach as the Fastspeech2 architecture \cite{ren2020fastspeech}, additional modules such as duration predictor and pitch predictor modules can be introduced to the system. These modules can be trained jointly or independently. The addition of duration and pitch predictors can be advantageous: it would provide explicit control over speaking rate and pitch. This would allow for real-time manipulation of these aspects during intelligibility experiments, enabling researchers to systematically modify speaking rate and pitch in non-native speech and investigate their impact on comprehensibility. By exploring the optimal settings that maximize intelligibility, this research direction may contribute to a more comprehensive understanding of the role of speaking rate and pitch in non-native speech perception and inform future studies to improve self-imitation tools in computer assisted pronunciation training.

Another avenue for improvement is to use HuBERT \cite{hsu2021hubert} codes as an alternative means of speech representation. In contrast to PPGs, HuBERT captures acoustic and linguistic features by leveraging self-supervised learning. Unlike traditional methods that rely on handcrafted features or spectrograms, HuBERT learns directly from the raw audio signal. This approach allows it to extract intricate details and capture contextual information, leading to more robust and accurate speech processing. By integrating HuBERT into our system, we can potentially benefit from improved speech representation, enhanced naturalness in speech generation, and thus, increased overall system performance.

The proposed system has potential applications in emotion transfer VC tasks \cite{he2021improved}, where the goal is to alter the emotional content of a given speech signal while preserving other characteristics such as speaker identity. An interesting direction for future work is to assess the system's performance in these tasks and evaluate its effectiveness in conveying and transforming emotions. To begin exploring emotion transfer, one can initiate the process by fine-tuning the system on a corpus consisting of recordings that encompass a wide range of emotions. By exposing the model to diverse emotional expressions, it can learn to capture and understand the underlying patterns and nuances associated with different emotional states. Fine-tuning on such a dataset would enable the system to acquire a more comprehensive understanding of emotional speech and enhance its ability to accurately convey and transform emotions.
\vspace{-2mm}
\subsection{Comprehensibility study}
We used our proposed accent conversion model to investigate the relative contributions of voice quality, segmental, and prosodic features on the perception of comprehensibility in non-native speech. Our findings align with previous research indicating that mimicking the pronunciation patterns of native speakers can enhance the comprehensibility of non-native speech. However, our study offers a unique contribution by specifically quantifying the individual effects of segmental and prosodic features on the comprehensibility of non-native speech and demonstrating the effectiveness of modifying these features to match those of native speakers. Interestingly, our results indicate that segmental features have a larger influence on comprehensibility compared to prosodic features. This implies that in endeavors aimed at enhancing the comprehensibility of non-native speakers, particular attention should be given to the accuracy of their segmental attributes, such as consonant and vowel production, rather than focusing solely on prosodic features like intonation and stress. These findings differ from prior studies on non-native speech that suggest prosody is more important than segmental features \cite{anderson1992relationship,munro1995foreign}. A potential explanation for this discrepancy is that our study used short utterances ($\approx4$ seconds) of \textit{read} speech. Further investigations incorporating spontaneous speech and longer utterances (reading comprehension tasks or conversational speech) could offer a more comprehensive understanding of the interplay between segmental and prosodic attributes in non-native speech perception. Additionally, introducing variability in semantics and examining how intelligibility is affected by predictable versus unpredictable semantic contexts would provide further insights into the relationship between context and speech perception.

Our study has several limitations. The sample size was relatively small and may not be representative of the larger population of non-native speakers. Future research with larger and more diverse participant groups could provide a broader perspective on the effects observed in our study. Our study focused on listeners who were native speakers of American English, and we did not investigate the influence of individual differences in listener characteristics. Exploring the impact of listener factors (e.g., language proficiency, cultural background and exposure to non-native speech) on comprehensibility judgments would be valuable in understanding the variability in perception and designing targeted accent reduction programs for specific listener profiles. Furthermore, future research could examine how various speech attributes affect comprehensibility in non-native listeners. Investigating the listener adaptation process \cite{xie2018rapid} and the dynamic nature of speech perception would shed light on the underlying mechanisms involved in intercultural communication.

Our methodology has potential applications in the field of socio-phonetics, offering opportunities to investigate the influence of non-native accents on social biases, particularly in employment and professional contexts. This research can provide valuable insights that inform policies aimed at reducing discrimination based on non-native accents and dialects. For example, to further explore the impact of non-native accents on credibility, our methodology can be combined with experiments similar to those conducted by Lev-Ari \textit{et al.} \cite{lev2010don}. In their study, participants were presented with trivia statements spoken by speakers with varying degrees of accentedness and were asked to judge the truthfulness of the statements. Future studies can adopt a similar approach to isolate and quantify the specific effects of segmental and prosodic features arising from non-native accents, providing an understanding of their influence on perceived credibility. By addressing these research areas, future studies can advance our understanding of non-native speech intelligibility and contribute to effective interventions and strategies for improving cross-cultural communication.

\section{Conclusion}
We have introduced an approach to accent conversion that overcomes the limitations of conventional methods by allowing independent control of segmental and prosodic characteristics in the resynthesis of L2 speech. Our proposed model uses a discretization step, involving subsampling in time and vector quantization of phonetic content, to facilitate accurate prosody transfer. Objective and subjective experiments indicate that the discretization step (i.e., vector quantization with duplicate removal) is critical to achieve successful prosody transfer, and that vector quantization improves the preservation of speaker identity during the conversion process.

By leveraging the capability to control segmental and prosodic characteristics independently, we were able to investigate their relative impact on the comprehensibility of non-native speech. Our findings revealed that segmental features exerted a stronger influence on speech comprehensibility compared to prosodic characteristics. It is worth noting that these results were obtained on a corpus of \textit{read} speech, which lacks the rich prosody of conversational speech. Future studies could consider incorporating a more diverse range of speech styles and explore the effects on social biases related to accent perception. The insights gained from these future investigations can inform the development of targeted interventions in pronunciation training to enhance overall comprehensibility and address potential biases in spoken communication.


\bibliographystyle{IEEEtran}
\bibliography{mybib}

\vfill

\end{document}